\documentclass[sigconf]{acmart}

\AtBeginDocument{%
  \providecommand\BibTeX{{%
    \normalfont B\kern-0.5em{\scshape i\kern-0.25em b}\kern-0.8em\TeX}}}

\copyrightyear{2023}
\acmYear{2023}
\setcopyright{acmlicensed}
\acmConference[KDD '23] {Proceedings of the 29th ACM SIGKDD Conference on Knowledge Discovery and Data Mining}{August 6--10, 2023}{Long Beach, CA, USA.}
\acmBooktitle{Proceedings of the 29th ACM SIGKDD Conference on Knowledge Discovery and Data Mining (KDD '23), August 6--10, 2023, Long Beach, CA, USA}
\acmPrice{15.00}
\acmISBN{979-8-4007-0103-0/23/08}
\acmDOI{10.1145/3580305.3599802}

\begin{document}

\title{AlerTiger: Deep Learning for AI Model Health Monitoring at LinkedIn}

\author{Zhentao Xu}
\affiliation{%
  \institution{LinkedIn Corporation}
  \city{Sunnyvale}
  \state{CA}
  \country{USA}
  \postcode{94085}
}
\email{zhexu@linkedin.com}

\author{Ruoying Wang}
\affiliation{%
  \institution{LinkedIn Corporation}
  \city{Sunnyvale}
  \state{CA}
  \country{USA}
  \postcode{94085}
}
\email{ruowang@linkedin.com}

\author{Girish Balaji}
\affiliation{%
  \institution{LinkedIn Corporation}
  \city{Sunnyvale}
  \state{CA}
  \country{USA}
  \postcode{94085}
}
\email{gbalaji@linkedin.com}

\author{Manas Bundele}
\affiliation{%
  \institution{LinkedIn Corporation}
  \city{Sunnyvale}
  \state{CA}
  \country{USA}
  \postcode{94085}
}
\email{mbundele@linkedin.com}
 
\author{Xiaofei Liu}
\affiliation{%
  \institution{LinkedIn Corporation}
  \city{Sunnyvale}
  \state{CA}
  \country{USA}
  \postcode{94085}
}
\email{xiaoliu@linkedin.com}

\author{Leo Liu}
\affiliation{%
  \institution{LinkedIn Corporation}
  \city{Sunnyvale}
  \state{CA}
  \country{USA}
  \postcode{94085}
}
\email{leoliu@linkedin.com}

\author{Tie Wang}
\affiliation{%
  \institution{LinkedIn Corporation}
  \city{Sunnyvale}
  \state{CA}
  \country{USA}
  \postcode{94085}
}
\email{tiewang@linkedin.com}

\renewcommand{\shortauthors}{Zhentao Xu et al.}

\begin{abstract}
Data-driven companies use AI models extensively to develop products and intelligent business solutions, making the health of these models crucial for business success. Model monitoring and alerting in industries pose unique challenges, including a lack of clear model health metrics definition, label sparsity, and fast model iterations that result in short-lived models and features. As a product, there are also requirements for scalability, generalizability, and explainability. To tackle these challenges, we propose AlerTiger, a deep-learning-based MLOps model monitoring system that helps AI teams across the company monitor their AI models' health by detecting anomalies in models' input features and output score over time. The system consists of four major steps: model statistics generation, deep-learning-based anomaly detection, anomaly post-processing, and user alerting. Our solution generates three categories of statistics to indicate AI model health, offers a two-stage deep anomaly detection solution to address label sparsity and attain the generalizability of monitoring new models, and provides holistic reports for actionable alerts. This approach has been deployed to most of LinkedIn's production AI models for over a year and has identified several model issues that later led to significant business metric gains after fixing.
\end{abstract}





\maketitle

\section{Introduction}
With the development of a large number of diverse AI-driven products in enterprises, the field of ML Operations (MLOps) has recently grown in interest both in academic and industrial settings \cite{tamburri2020sustainable}. MLOps describes the processes by which data-driven models underlying an enterprise's applications are developed, deployed, and maintained \cite{testi2022mlops}. Companies like LinkedIn and Google have developed MLOps platforms that handle the “end to end” machine learning lifecycle with the following primary benefits: efficient model development and productionization in applications, scalability concerning managing thousands of different models, and risk reduction by enabling model observability and alerting \cite{testi2022mlops}. 

At LinkedIn, the Health Assurance component of the ProML platform \cite{linkedin2021health} handles the maintainability aspect of MLOps. This component monitors all models' input features and output scores  during online inference and curates daily statistics like mean or missing percentages on all these entities. In LinkedIn's MLOps platform, every deployed model has a set lifespan controlled by each ML Engineering team which controls the ramp-up and ramp-down of production traffic routed toward this model. If we identify sufficiently concerning behavior in a model's features or score on any given day, the MLOps platform sends a single automatic alert to the team owning the model.

To provide ML engineers confidence in their deployed models, we need a method of automatically alerting them whenever we can identify concerning behavior. An anomalous period of a time series refers to a time range during which its values deviate significantly from its historical patterns. During day-to-day model serving, anomalies in the feature and score statistics can often lead to detrimental model performance and, consequently, business impact. Engineers use these alerts to identify and fix issues, for example, fixing broken feature processing jobs \cite{laptev2015generic} or retraining models.

There are many challenges in designing an industrial model health monitoring solution.

\textbf{Challenge 1: Lack of Unified Definition of Model Health.} While the software development community has long accepted latency and CPU utilization as well-measured system health indicators, there is no consensus on a unified measure of model health. ML engineers, in many cases, deal with model health degradation reactively only after the impact on business metrics materializes. That is when an  issue identification and triaging session will be opened. We handle this challenge using a bottom-up approach by generating  three categories of statistics to indicate AI model healthiness (Section \ref{subsection:Model Health Statistics Generation}) and alerting on time series deviations for all input features and output scores (Section \ref{subsection:Univariate Deep Learning Anomaly Detection}). We then combine such anomaly patterns with other metadata on traffic and importance to decide whether we should notify the ML engineers of this abnormal observation, preemptively (Section \ref{subsection:Anomaly Post Processing}). 

\textbf{Challenge 2: Anomaly Labels Sparsity.} Unsupervised methods could introduce a lot of noisy alerts. We handle data sparsity by maintaining an MLOps training dataset that combines model owners and domain expert labels on unlabeled data based on our anomaly pattern definition. Furthermore, our two-stage deep learning model with anomaly post-processing significantly reduces noisy alerts (Section \ref{subsubsection:Model Structure}).

\textbf{Challenge 3: Generalization.} Traditionally, time series will be fit on-the-go on a given length of historical data. This process prevented the possibility of complex models, and model performance was tightly bound to the quality of historical data. In our solution, we train one model offline with different patterned data (e.g., spiky time series, time series with or without seasonality, as detailed in Section \ref{subsubsection:Anomaly Pattern Definition}) and then serve online. This makes our model suitable for various time series without fine-tuning online, leading to low onboarding costs and wide applicability at LinkedIn.

\textbf{Challenge 4: Explainability.} As a product, we must balance solution complexity and ease of understanding, leading to actionable and accurate insights into model health. Simply applying an anomaly detection algorithm on one statistic will not be sufficient for delivering actionable alerts to users. In our solution, we detect anomalies on various model health statistics and combine the detection result into a holistic report to AI model owners, highlighting the interaction between anomaly dimensions (Section \ref{subsubsection:Descriptive Alert Report}).

To tackle the aforementioned problems, we aim to develop an anomaly detection method for monitoring the AI model's health that is accurate, efficient, and generalizable. In this paper, we propose an end-to-end AI model healthiness monitoring solution, named AlerTiger, with four major steps: (1)  AI model healthiness statistics generation; (2) supervised univariate deep-learning-based time series anomaly detection for detecting anomalies on individual feature statistics separately; (3) anomaly grouping logic to aggregate univariate time series to the model level, and anomaly filtering criteria for high precision; (4) the anomaly information are combined into an alert report for issue investigation and fixing. The code of AlerTiger is available on LinkedIn's GitHub page\footnote{\url{https://github.com/linkedin/AlerTiger}}.

As shown in the experiment result (Section \ref{section:Experiments}), our proposed algorithm is more accurate than traditional unsupervised time series anomaly detection algorithms and supervised deep learning forecasting-based anomaly detection algorithms. The anomaly filtering post processing step is effective in keeping anomalies that are severe and worth investigating. AlerTiger's serving time is 2X faster than competitive algorithms, while training is manageable on a single-CPU machine within hours, making the solution practical for productionization in the industry.

This paper's \textbf{contributions} are highlighted below:
\begin{itemize}
    \item We build a novel, end-end, data-driven solution for monitoring AI models in the industrial setting by clearly defining model healthiness with concerning patterns in critical features' statistics. We also demonstrate how the anomalies in feature distribution statistics are closely related to common AI model issues with experiments.
    \item To the best of our knowledge, this is among the earilest attempts of applying deep-learning-based forecasting and anomaly detection algorithms in the AI model monitoring use case. These algorithms achieve high detection performance in LinkedIn's real production data. The two-stage deep model architecture helps achieve high performance from supervised learning while requiring low labeling efforts.
    \item By adding quantile loss to the commonly used RMSE, our neural network estimates a non-parametric distribution of the underlying data, which is practical when real data usually does not follow Gaussian distribution. 
    \item In the production MLOps model alerting setting, our solution demonstrates broad generalizability. We successfully trained one model capable of accurately detecting anomalies for virtually all different AI features with different historical patterns. We have productionized the solution for most of LinkedIn's AI models with high performance. 
\end{itemize}

The rest of this paper is organized as follows. In Section \ref{section:Related Work}, we describe the research and industrial practices in MLOps and time series anomaly detection fields. In Section \ref{section:Methodology}, we discuss the model monitoring product and the deep-learning-based model that powers this system. We further compare our system with different state-of-the-art techniques in the Section \ref{section:Experiments}. Finally, we conclude our work and present future work in the Section \ref{section:Conclusion}.

\section{Related Work}
\label{section:Related Work}

\subsection{ML Operations (MLOps)}
\label{subsection:ML Operations (MLOps)}

In recent years, machine learning has evolved from a research area into productionized solutions for business problems \cite{paleyes2022challenges}. Several AI-focused companies have built standardized MLOps platforms to speed up the machine learning algorithm development process and support model deployment and maintenance at scale. Notable platforms include TensorFlow Extended (TFX) used at Google \cite{baylor2017tfx}, SageMaker for AWS \cite{nigenda2021amazon}, Azure Machine Learning \cite{dem1082023mlops}, and Michaelangelo at Uber \cite{uber2017meet}. The Pro-ML platform is used at LinkedIn \cite{joel2019scaling}. As an indispensable part of the end-to-end MLOps system, model monitoring takes a data-centric approach to ensure the sanity of the petabytes of data flowing in the system \cite{breck2019data}. As claimed by Databricks, $15\%$ of machine learning projects fail due to fragile ML systems, including unstable online features; further, a large commonly-overlooked cost is managing the health of deployed models in production and accurately monitoring its critical metrics to demonstrate the model's value \cite{databricks2021machine}. 

To tackle the MLOps monitoring problem, the first step is to monitor the data that are clear indicators of problems in the model deployment lifecycle. On the model input side, TFX deploys a data validation system \cite{breck2019data} that checks schema correctness and ensures that training code co-evolves with data schema changes. Arize AI \cite{arizeai2021machine} takes the monitoring definition one step further by monitoring  the symptoms of model degradation, including concept and feature drift, online and offline inconsistency, and outliers in the feature distribution. AWS's SageMaker \cite{nigenda2021amazon} monitors four aspects of MLOps, including data quality, model quality, bias drift, and feature attribution drift. Our AlerTiger system provides an end-end model monitoring solution with model healthiness statistics collection, deep-learning-based anomaly detection on AI models' input features and output scores over time, and an actionable alerting solution.

With the monitoring data, surfacing the right problem at the right time becomes the biggest challenge. It is unscalable to have users set manual rules, and in many cases, users are interested in nuanced feature distribution changes that are hard to be described as rules. This is where machine learning becomes helpful in surfacing and alerting users on abnormal data patterns. The anomaly detection algorithm is at the core of our AlerTiger system.

\subsection{Time Series Anomaly Detection}
\label{subsection:Time Series Anomaly Detection}

For anomaly detection, a general framework consists of 3 steps \cite{wang2020deep}: 
\begin{itemize}
    \item Step-1: learn the mapping of data representation $f(·; \theta): x \rightarrow y$, where $x$ and $y$ represent the input features and the output prediction, respectively. $\theta$ is the weight to be learned and common elements include mean, std, etc., of the next data points. 
    \item Step-2:  learn the anomaly measurement $d(f(x); \eta)$, where $\eta$ is the weight to be learned.
    \item Step-3:  define a threshold $\varsigma$ to identify the anomalies. 
\end{itemize}

The purpose of the first step, in the time series context, is to remove any recurring seasonality or time dependencies. So that in the second step, we are left with more uniform and independently distributed errors (difference between the observed and the predicted from step 1). There is a wide array of candidate time series prediction algorithms one can use for the first step mapping function. Traditional statistical models run regressions on time and seasonality, where the model structure can be Spline Regression \cite{marsh2001spline}, Bayesian Structural Time Series (BSTS \cite{scott2014predicting}), etc.  Alternatively, one can run regression on the time dependencies through the autoregressive-moving-average (ARMA) model and its variations such as ARIMA and ARIMAX \cite{yu2016improved}. In recent years, deep learning networks have also been applied to the first step and showed performance lift. Common networks include LSTM \cite{salinas2020deepar, ergen2019unsupervised, malhotra2015long} and CNN \cite{munir2018deepant}. This mapping function learning step is usually a supervised regression model, with RMSE as the loss function or employing a maximum likelihood estimate. Some models are also structured to output the distribution parameters, which can be leveraged by the second step of anomaly detection. Statistical models, for example, can estimate the standard deviation of a Gaussian distribution with the residuals, which was deployed as one of the anomaly detection algorithms in AWS's SageMaker \cite{nigenda2021amazon}. \cite{salinas2020deepar} parameterized the distribution with Gaussian and Negative Binomial. Our model, on the other hand, takes a non-parametric approach where in addition to mean prediction, we add a quantile loss to our loss function to estimate the upper and lower bounds given the significance level we are interested in. The benefit of this approach is we do not need to make any parametric assumption of the residuals so that the method is easier to apply to irregular time series that do not follow common distributions in practice. 

The second and third steps of time series anomaly detection are setting an anomaly score for any given data point and determining an anomaly threshold for decision-making. Due to the label sparsity nature of anomaly detection, unsupervised learning techniques are usually used. A commonly used approach in the industry is calculating the z-score of the first-step prediction residual. Then one can apply the 3-sigma approach or set a p-value threshold (\cite{malhotra2015long}, \cite{shipmon2017time} for Google, \cite{laptev2015generic} for Yahoo, \cite{nigenda2021amazon} for AWS). Alternatively, one can map the residuals into kernel space and learn the hyperplane so that most of the data points in training lie on one side, and anything on the other side can be viewed as an anomaly \cite{scholkopf2000support, ergen2019unsupervised}. In practice, however, unsupervised anomaly detectors, due to their probabilistic nature, bring in random noises, which, when accumulated, could lead to many false alerts. \cite{hundman2018detecting} proposes to use dynamic thresholding, which leverages historical anomalies. But this method requires long history input and poses latency concerns for real-time applications. \cite{xu2018unsupervised} proposes Donut, a framework that uses the variational autoencoder (VAE) \cite{kingma2013auto} to estimate the first and second steps together. However, it is still subject to the common problems with unsupervised learning. Moreover, the training and deployment cost of such a model is high \cite{ren2019time}. In our framework, we take a supervised approach to leverage as much data as possible. Since we have a clear anomaly pattern definition, it is easier for us to gather labels for training, and the model trained with this data is more catered to our ML system use case. Over time, as we accumulate more labels, the model can be retrained and updated.

Another common scenario in time series anomaly detection is the multivariate case. They are usually applied in IoT \cite{munir2018deepant, gugulothu2018sparse,zhang2019deep}, spacecraft \cite{hundman2018detecting}, and cyber attacks \cite{filonov2016multivariate}, where recurrent neural networks \cite{munir2018deepant, gugulothu2018sparse,filonov2016multivariate} and convolutional neural networks \cite{zhang2019deep} are commonly used. Dimension reduction is at the core of the main contributions of the deep learning network \cite{gugulothu2018sparse}. These approaches, however, are not directly applicable to the model monitoring case. This is predominantly due to the varying dimensions of input and output statistics (termed "sensors") across different AI models. As a consequence, each of the thousands of rapidly iterating AI models to be monitored in the system necessitates its own anomaly detection model. This poses a significant challenge in terms of onboarding costs, generalization, and infrastructure functionality within the industrial setting. We currently approach the problem by applying the AlerTiger univariate time series anomaly detection algorithm on each univariate model healthiness statistics time series. Then apply anomaly grouping to decide whether to surface anomalies at the model level.

\section{Methodology}
\label{section:Methodology}

\begin{figure*}[hbt!]
    \centering
    \includegraphics[width=\textwidth]{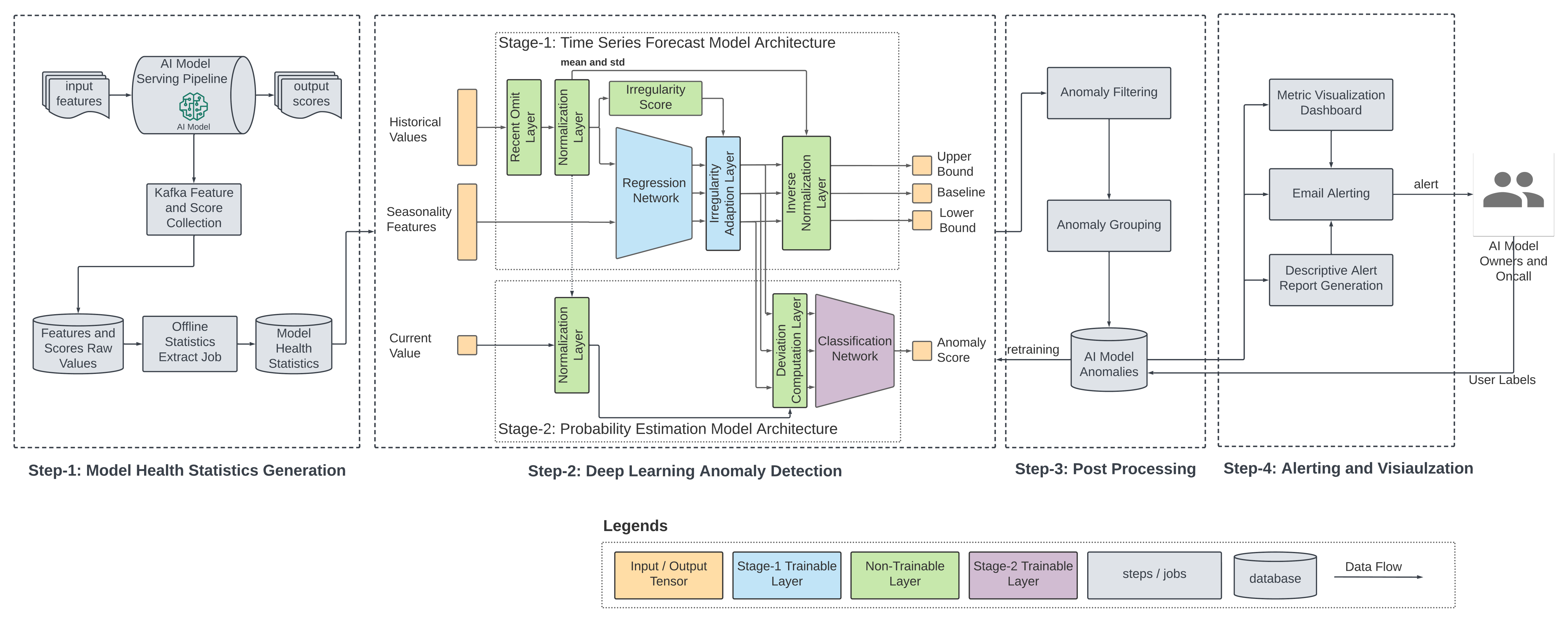}
    \caption{Architecture diagram for the AlerTiger AI model monitoring system}
    \label{fig:system diagram}
\end{figure*} 

In this section, we introduce and describe the AlerTiger system for detecting the AI models' anomalies. Figure \ref{fig:system diagram} shows the end-to-end system for model health monitoring, which consists of four major components:

\begin{itemize}
    \item Model Health Statistics Generation: during model scoring, the serving pipeline will emit and store them to HDFS via Kafka. A daily offline Spark job then aggregates the data for each model and computes the daily healthiness statistics accordingly. 
    \item Deep Learning Anomaly Detection: We run a deep learning anomaly detection algorithm on the healthiness statistics of a model.
    \item Anomaly Post Processing: We apply anomaly filtering and grouping logic, eventually getting the anomaly at the model level. 
    \item Alerting and Visualization: The model health statistics and information are compiled into a holistic report and sent to AI model owners through emails.
\end{itemize}

\subsection{Model Health Statistics Generation}
\label{subsection:Model Health Statistics Generation}

We generate three categories of statistics to indicate AI model healthiness: model input feature statistics, model output score statistics, and auxiliary metadata.

Model input feature statistics refers to the summary statistics of the model's input features. At LinkedIn, we monitor individual features' mean, standard deviation, and quantiles aggregated daily. Non-numeric features such as categorical features and embedding features can be considered as a collection of numeric values where the statistics can be extracted from each category or dimension respectively. We also monitor feature coverage, i.e., the percentage of non-default values (e.g., non-zeros) and non-missing values, as in many cases, fluctuations in the proportion of zero and the proportion of nulls are key indicators of feature health problems, such as upstream raw data missing, data schema change or feature generation pipeline failure, that may otherwise easily get unnoticed.

Model output statistics include model daily output scores distribution (e.g., mean, standard deviation, quantiles). We also monitor model performance statistics such as the Observed to Expected ratio (O/E)  and normalized discounted cumulative gain (NDCG) to reflect model performance since they are usually considered key performance indicators for models and are crucial for AI engineers.

Auxiliary metadata captures information about the model. We track daily model traffic (i.e., the number of scoring requests within a day), model traffic ratio (i.e., the percentage of the  model's traffic among a group of models behind a product, as there can be multiple versions of models running simutaneously  for experimentation or A/B testing), and feature importance score. As we will discuss in the Section \ref{subsubsection:Anomaly Filters}, the meta information can effectively filters out less important alerts and provides key information in the alert report for AI model owners.

In the rest of the paper, we will use $T_{m,f,s}(t)$, or $T(t)$ in short, to denote a univariate time series value for a particular model $m$, input feature (or output score) $f$, statistic $s$, at the timestamp $t$.

\subsection{Pointwise Anomaly Detection}
\label{subsection:Univariate Deep Learning Anomaly Detection}

As discussed in Section \ref{subsection:Model Health Statistics Generation}, we consider various statistics such as mean, percentiles, and missing percentage to describe the status of AI models, each is represented as a univariate time series $T(t)$ . In this section, we consider each univariate time series separately and detect the issues for them. We describe our model in the context of daily granularity data.

\subsubsection{Feature Engineering}
\label{subsubsection:Feature Engineering}

The inputs of the univariate deep learning model contain three features: the historical time series vector $ \bar{h}:=T(t-H:t-1) \in \mathbb{R}^H$ presenting the value of the statistics during the historical window of length $H$; the day-of-week one-hot-encoding feature $\bar{s} \in \mathbb{R}^7$; and the latest observed value $c :=T(t) \in \mathbb{R}$. We chose the day-of-week seasonality feature because daily granularity statistics usually show weekly pattern. The historical length $H$ can be chosen based on AI models' common lifespan length and the tradeoff between detection performance and detection cold-start period length: A longer $H$ usually leads to better detection performance; while at the same time increasing the cold-start period length, during which a model is not covered by alerting. At LinkedIn, we constructed a short-term anomaly detection algorithm with $H=14$ days for monitoring newly onboarded models within the past 2 to 4 weeks and a long-term algorithm with $H=28$ days for monitoring models longer than 4 weeks. 

\subsubsection{Model Structure}
\label{subsubsection:Model Structure}
As shown in Figure \ref{fig:system diagram} second block, the AlerTiger model contains two stages. The first one-step forecasting stage takes as inputs the historical time series vector $\bar{h}\in \mathbb{R}^H$, and the day-of-week one-hot-encoding feature $\bar{s}\in \mathbb{R}^7$, and generates the expected value $\hat{T}(t)$ and confidence boundaries $\hat{T}_{l}(t)$ and $\hat{T}_{u}(t)$ at the timestamp $t$. The second anomaly classification stage uses the prediction result $\hat{T}(t)$,  $\hat{T}_{l}(t)$ and $\hat{T}_{u}(t)$ from the forecasting stage and the current observed value $T(t)$ as inputs, both in the normalized space, to generate the anomaly probability. 

In the first stage (time series forecast stage), historical time series feature $\bar{h}$ is first pre-processed by the Recent Omit Layer and the Layer Normalization Layer (see section \ref{paragraph:Generalizability by Layer Normalization}). The pre-processed historical data and day-of-week seasonality features are concatenated and fed into a regression network to forecast the expected baseline value and boundaries. The regression network is generic and can be any network architecture, such as a convolutional neural network (CNN), long-short-term memory (LSTM), or variational autoencoder (VAE). For the sake of simplicity, here we picked the multi-layer perceptron (MLP) network as the regression network. As shown in the loss function below, the first stage's loss is a combination of mean squared error (MSE) on the forecasted baseline and quantile loss on lower and upper boundaries. Note that the we only consider the non-anomaly data points in the mean square error.

\[L_{MSE} := \frac{1}{N}\sum_{t=1}^{N}{[T(t) - \hat{T}(t)]^2}\]
\[L_{l} := \frac{1}{N}\sum_{t=1}^{N}{ \max{\{\tau_{l}\cdot[T(t)-\hat{T}_{l}(t)], (\tau_{l}-1)\cdot[T(t)-\hat{T}_{l}(t)]\}}}\]
\[L_{u} := \frac{1}{N}\sum_{t=1}^{N}{ \max{\{\tau_{u}\cdot[T(t)-\hat{T}_{u}(t)], (\tau_{u}-1)\cdot[T(t)-\hat{T}_{u}(t)]\}}}\]
\[L_{forecast}:=L_{MSE} + \lambda\cdot (L_{l} + L_{u})\]
where $N$ is the number of training data points, $L_{MSE}$ is forecasted baseline error, $L_l$ and $L_u$ are quantile losses, $\tau_l$ and $\tau_u$ are corresponding quantiles, $\lambda$ is the relative weights of quantile losses. The hyper-parameters are tuned on validation dataset and we use $\tau_l=2.5\%$, $\tau_u=97.5\%$, and $\lambda = 1$.

The second stage (anomaly probability estimation) estimates the anomaly probability using the output from the first stage $\hat{T}(t)$, $\hat{T}_{l}(t)$, $\hat{T}_{u}(t)$ and the observed value $T(t)$. To achieve this, the model first calculates the deviations between the observed and the three forecasted value, then feed the deviations into a classification network which outputs the probability of $T(t)$ being abnormal.

The final step in anomaly detection is anomaly classification. A naive way of anomaly detection is simply classifying out-of-boundary observations as anomalies. However, based on our experiments, such method usually led to a high false alert rate. We jointly consider the out-of-boundary observation and anomaly probability score in our deep learning model, i.e., $\hat{Y}(t)=I(T(t) \notin [\hat{T}_{l}(t), \hat{T}_{u}(t)]) \cap I(P_{anomaly} \geq \theta_{anomaly})$, where $\hat{Y}(t)$ is the predicted anomaly, $I(\cdot)$ is indicator function, and $\theta_{anomaly}$ is the anomaly score threshold which can be tuned on validation dataset, we use $\theta_{anomaly} = 0.2$.

\subsubsection{Model Details}
\label{subsubsection:Model Details}
The remaining part of this section will focus on explaining some technical details/improvements to improve the detection quality of the deep learning model.

\paragraph{Generalizability by Layer Normalization} 
\label{paragraph:Generalizability by Layer Normalization}
Time series problems usually normalize global time series by normalizing the entire time series during training. However, this mechanism has two risks: (1) information leakage: during the normalization, mean and standard deviation are derived from the entire time series, meaning future statistics values can impact past time series values in the normalized space. (2) over-complicated input space for model training: even historical time series having the same shapes might be mapped to different data points after normalization and thus have different prediction values.

Layer normalization was originally proposed by \cite{ba2016layer}, which transforms the inputs to have zero mean and unit variance within each feature. We apply layer normalization to the raw historical input time series feature vector $\bar{h}$; we use the standard zero-mean unit-standard deviation normalization method here. The normalization parameter (mean, standard deviation) is directly sent to the inverse normalization layer before the time series predictor's output, including predicted baseline and boundaries. The two aforementioned risks can be solved with layer normalization: the information leakage problem is resolved because each historical feature vector will be equally standardized to zero mean and unit standard deviation, disregarding the impact from future time series values. The model's input space is simplified because all input historical value features are mapped to the same scales, improving the generalizability of the deep model to time series of various types.

\paragraph{Robustness by Omitting Recent Values}
\label{paragraph:Robustness by Omitting Recent Values}
One typical issue we observed in training the prediction model is over-adaptation: a higher adaptation means a shorter time to adapt the prediction to new trends. Adaptation may be good as it can avoid sending repeated alerts to users; however, there are two severe issues with over-adaptation: (1) the deviation between prediction and observation is usually too small to trigger an alert. (2) hard to estimate the actual anomaly duration, which is one key attribute for determining anomaly significance. We improve the network's robustness against anomalies not yet labeled by users by skipping the most recent $K$ days' data points in the time series input. The advantage is to avoid the network over-emphasizing the most recent data over past data. In our use case, we use $K=3$ days, which is a good balance between performance and adaptation.

\paragraph{Irregularity Estimation on Limited Data}
\label{paragraph:Irregularity  Estimation on Limited Data}
Another observation of the above deep model is that irregular statistics usually have unacceptably higher alert rates than regular statistics. This unbalanced alert rate behavior is understandable but not desirable. On the one hand, it is understandable because irregular statistics are naturally unpredictable; on the other hand, it is undesirable because sending AI model owners a flood of alert emails due to the irregular statistics will cause fatigue in model owners.

We use an irregularity score estimation mechanism to overcome the above issue. As the time series historical length for model healthiness monitoring is usually short, we use the interquartile range (IQR) of the week-over-week difference $\bar{\delta} \in \mathbb{R}^{H-7}$ to indicate the "irregularity" $S_{irreg}$ of the time series. To reduce bias, we propose a trainable irregularity adaption layer to transform this score so that we can eventually use this score to amplify the forecast boundaries for getting the final boundaries $\hat{T}_{l}(t)$  and $\hat{T}_{u}(t)$. In our use case, the irregularity adaptation layer is realized as an MLP layer with a sigmoid activation function.

\[\bar{\delta} := T(t-H: t-8) -  T(t-H+7: t-1) \in \mathbb{R}^{H-7} \]
\[S_{irreg} := MLP(IQR(\bar{\delta}))\]
\[\hat{T}_{l}(t) := S_{irreg}\cdot( \hat{T}_{l}(t) - \hat{T}(t) ) +\hat{T}(t)\]
\[\hat{T}_{u}(t) := S_{irreg}\cdot( \hat{T}_{u}(t) - \hat{T}(t) ) +\hat{T}(t)\]

The intuition behind the irregularity estimation is that a regular time series with strong weekly pattern leads to similar week-over-week differences $\delta_i=T(i-7) - T(i)$, thus having a smaller IQR; in constrast, an irregular time series without seasonality patterns has variety of weekly differences, thus higher IQR. this estimation is robust against both the time series trends and the noisy spikes because IQR is a robust statistic. The score is insensitive to IQR's quantile parameters; we used 25\% and 75\% as the quantiles for IQR.

\subsection{Anomaly Post Processing}
\label{subsection:Anomaly Post Processing}

As discussed in Section \ref{subsection:Univariate Deep Learning Anomaly Detection}, for each statistic of each feature in the AI model, the deep learning model classifies each timestamp to be anomalous or not based on whether the classification model's anomaly score is above the threshold and whether the observed value is out of boundary. This section will discuss how we move one step further by combining domain knowledge into anomaly filtering to improve detection accuracy and grouping the univariate anomaly to model level. Note that although our current model uses a global filtering threshold, the filter threshold can be further customized based on individual model owner's domain knowledge and requirements.

\subsubsection{Anomaly Filters}
\label{subsubsection:Anomaly Filters}

Anomaly filtering is essential for reducing false alerts and providing more relevant detection results to the users. We measure the four aspects of an anomaly for anomaly filtering: anomaly duration, severity, concurrent anomaly counts, and model traffic ratio, as discussed in the following subsections.

\paragraph{Anomaly Duration Filter}
\label{paragraph:Anomaly Duration Filter}
Anomaly duration represents how long an anomaly lasts. Quantitatively, it is measured by calculating the number of abnormal data points within the detection window $[t_s, t_e]$. We then apply threshold-based filtering to this anomaly duration. The intuition is that a long-lasting anomaly is usually more likely to be a real issue and have a continuous impact; in contrast, a single-point spike that gets recovered soon within one day may get relatively lower priority.  As an example, infra failure can be one possible reason for such short-term spikes; as infra issue has its DevOps alerting system, the issue can get fixed quickly, model owners may not worry much about such tentative anomalies. We use the duration threshold $\theta_{duration}=2$ days based on the above domain knowledge. 

\[\hat{Y}_{duration, [t_s, t_e]} := | \{t \in [t_s, t_e]  | \hat{Y}_t) \} | \geq \theta_{duration} \]

\paragraph{Anomaly Severity Filter}
\label{paragraph:Anomaly Severity Filter}

Anomaly severity represents how far the observed value deviated from predicted baseline. We define the anomaly severity by normalizing the absolute deviation $|\hat{T}(t)-T(t)|$ with prediction confidence $|\hat{T}_u(t) - \hat{T}_l(t)|$. The intuition behind this definition is given the same absolute deviation, a time series that the model is confident in the prediction result is more concerning to AI engineers than a noisy time series with large variance. We turned the severity threshold $\theta_{severity} = 1.3$ on validation dataset.
\[\hat{Y}_{severity} := \frac{|\hat{T}(t) - T(t)|}{|\hat{T}_u(t) - \hat{T}_l(t)|} \geq \theta_{severity}\]

\paragraph{Anomaly Concurrency Filter}
\label{paragraph:Anomaly Concurrency Filter}
Different statistics of the same feature, as well as different features and model scores  of an AI model can be correlated. In this case, when an issue happens, chances are that many statistics can be impacted simultaneously. A concurrency filter can effectively remove noisy alerts that may occur only on a small portion of statistics. In our default setting, we turned off the concurrency filter to ensure any statistics' anomaly be sent to model owners, while still provide the flexibility of updating this setting.
\[\hat{Y}_{concurrency} := \frac{\mathrm{\#\, abnormal\, statistics}}{\mathrm{\#\, statistics}} \geq \theta_{concurrency}\]

\paragraph{Model Traffic Ratio Filter}
\label{paragraph:Model Traffic Ratio Filter}
In the production there can be multiple versions of models running behind a product for experimentation or A/B testing. Not every model is equally important. For example, some AI teams may configure a very low traffic percentage only for testing a new model, where the model issue can happen but has a relatively lower impact on users. We define the model traffic ratio (MTR) to represent the percentage of traffic served by a specific model. For instance, on a specific timestamp, there are $N_{model}$ models serving a product, where the model $m$'s traffic is $C_m$, then we define the model traffic ratio following the formula and use threshold-based filtering on it. We use the model traffic ratio filter threshold $\theta_{MTR} = 3\%$ based on AI teams' domain knowledge.

\[\hat{Y}_{MTR} := \frac{C_m}{\sum_{m=1}^{N_{model}}{C_m} }\geq \theta_{MTR}\]

\subsubsection{Anomaly Grouping}
\label{subsubsection:Anomaly Grouping}
Since the anomaly detection is on a univariate time series for each feature-statistics or score-statistics pair, and we generate alert email notifications for each AI model, anomaly grouping is required for aggregating the univariate time series anomaly into the model level. 

We provide a web application where users can configure the alert dimensions (e.g., a subset of features, which statistics for each feature to alert), and the aggregation method. By default, we apply anomaly detection on all features for a model and monitor the feature-level mean and feature-level coverage statistics, together with the model score mean. We aggregate univariate anomalies to model-level using OR logic by default, i.e., any abnormal feature can trigger alert emails. However, we provide users with the option to select a subset of important features for monitoring with customized aggregation methods.

\subsection{Alerting and User Accountability}
\label{subsection:Alerting and User Accountability}

One goal of a model monitoring system in practice is that users fix the detected issues. To this end, we need to make our anomaly patterns easy to understand and provide enough information for AI engineers to debug. We handle this challenge by defining anomaly patterns and providing a holistic report for each abnormal models.

\subsubsection{Anomaly Pattern Definition}
\label{subsubsection:Anomaly Pattern Definition}

Anomaly pattern serves as a knowledge base for people to map the patterns to potential problems. Although time series anomaly definition can vary from one use case to another, we find two common anomaly patterns in model healthiness statistics: short-term spike anomaly pattern and long-term level-shift anomaly pattern. 
\begin{itemize}
    \item Short-term spike anomaly pattern: the statistic deviates from the normal range, then quickly recovers to the previous normal range within a short period (for example, three days)
    \item Long-term level-shift anomaly pattern: the statistic deviates from the normal range and stays in the abnormal new level.
\end{itemize}

\subsubsection{Descriptive Alert Report}
\label{subsubsection:Descriptive Alert Report}
\begin{figure}[hbt!]
    \centering
    \includegraphics[width=0.47\textwidth]{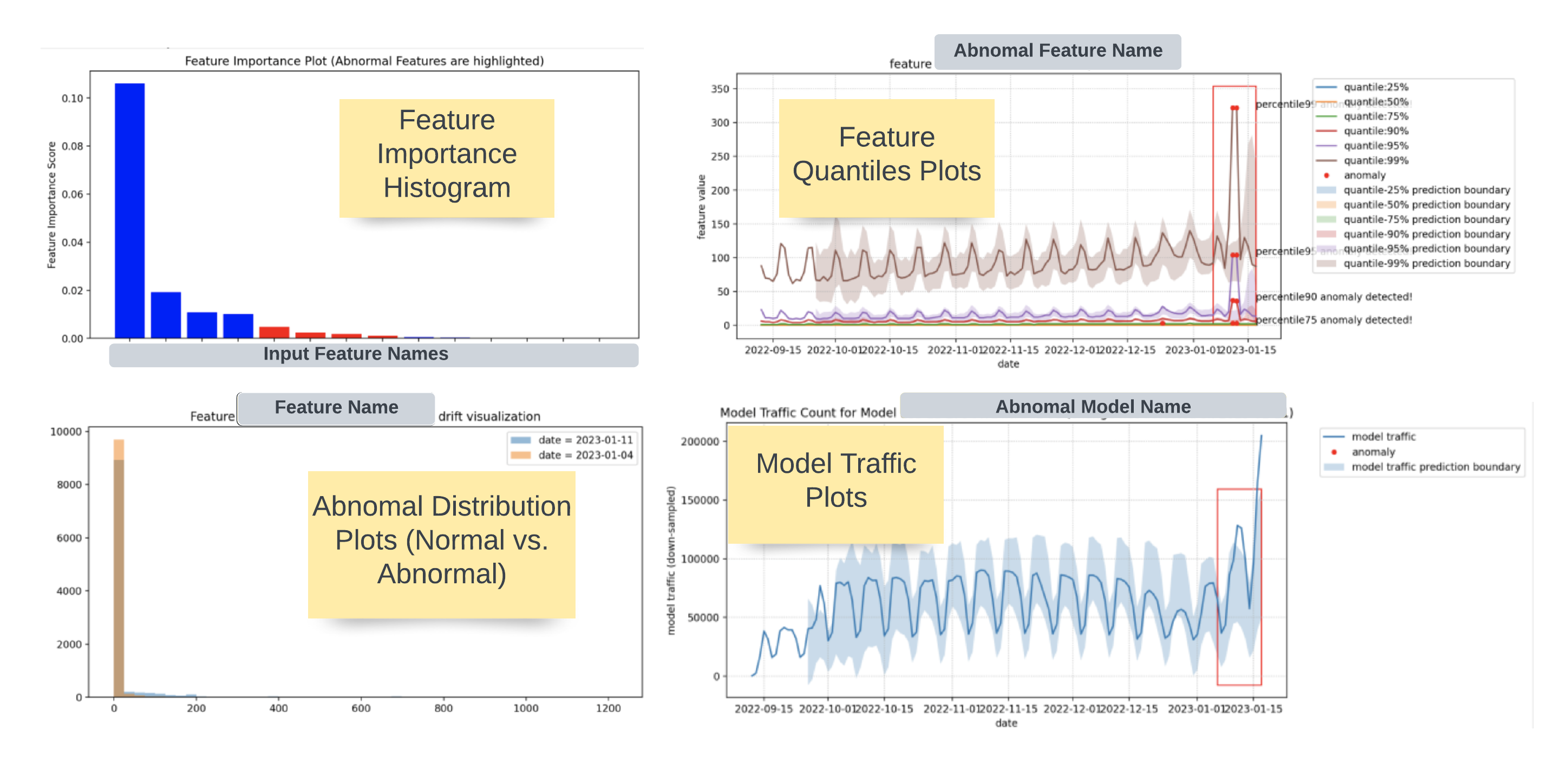}
    \caption{Descriptive Alert Report Screenshots. Visualizes the feature importance, quantiles, distribution and model traffic as examples, with sensitive information removed.}
    \label{fig:descriptive alert report}
\end{figure}

Industrial AI model systems are usually complex and a lack of holistic anomaly information of model is challenging for issue understanding and fixing. To overcome this challenge, we fetch the abnormal model's critical information that model owners care, including feature importance, example abnormal feature values, feature distribution, model traffic, anomaly patterns, etc.; we further compile the information into a holistic report with plots and descriptive information, explaining why we send alert and problematic aspects. The descriptive alert report has been launched for all AI teams during the past four months, and we saw a significant improvement on user response rate. Figure \ref{fig:descriptive alert report} shows some screenshots from an actual alert report.

\section{Experiments}
\label{section:Experiments}

\subsection{Datasets}
\label{subsection:Datasets}

\subsubsection{MLOps Model Health Data}
\label{subsubsection:MLOps Model Health Data}

\textbf{Training Dataset:} This dataset comprises curated daily time series and labels from three AI teams at LinkedIn. Model owners manually labeled each piece of univariate time series based on strict visual anomaly patterns definition described in Section \ref{subsubsection:Anomaly Pattern Definition}.  We used this dataset to train the two-stage univariate AlerTiger anomaly detection model.

\textbf{Evaluation Dataset:} This dataset comprises curated daily time series and labels collected across all AI teams onboarded to LinkedIn's MLOps platform. To evaluate our solution's real capability in identifying true production issue, we define the label at model level, where a positive label is defined as either model owners identified concerning behavior in model, or took actions in root-causing or fixing a real performance issues. This is stricter than the visual anomaly pattern definition we used in the training dataset. See Table \ref{tab:Dataset Summary Statistics} for summary statistics.

\subsubsection{Synthetic Evaluation Dataset}
\label{subsubsection:Synthetic Evaluation Dataset}
In addition to the MLOps model health data from LinkedIn, we create a synthetic time series dataset that can be used for extensive experimentation to reveal algorithms' performance.
We takes a two-step approach in synthetic data generation: (1) generate time series of various shapes with noises; (2) inject anomaly of various patterns, severity, and duration to time series.

To model the diverse time series types in actual production, we choose three time series shapes (sine wave, square wave, and constant value) with a fixed frequency and amplitude and varying noise added to it. Noise is generated by drawing random samples from a gaussian distribution with mean 0 and configurable standard deviation. We then injected anomalies of spike patterns or level-shift patterns, following the aforementioned anomaly pattern definition. The anomaly duration and intensity are configurable as we are interested in understanding how the model performs with various anomalies, and the anomaly location in the time series is stochastically chosen. For simplicity, we inject one anomaly per time series to help us understand the performance at various settings.

\begin{table}[]
    \centering
    \begin{tabular}{l|c|c|c}
     \hline
     Dataset Statistics & MLOps& MLOps & Synthetic  \\
     & Data for & Data for & Data for \\
         &  Training& Evaluation&  Evaluation \\\hline
         \# model &30+ & 1000+ & 101  \\
         \# features per model & 10+ & 10+ & 1.0 \\
         \# univariate time series & 300+ & 10000+ & 863\\
         \# anomaly intervals & 693 & 18 & 863\\
         \% abnormal days & 7.9\% & 0.14\% & 4.20\% \\\hline
    \end{tabular}
    \caption{Dataset Summary Statistics}
    \label{tab:Dataset Summary Statistics}
\end{table}


\subsection{Experiment Setups}
\label{subsection:Model Comparison}
We benchmark AlerTiger using the evaluation datasets in Section \ref{subsection:Datasets} against baseline algorithms Prophet, ARIMA, AR, DeepAR, and SARIMAX. We use the same "rolling window" setup to evaluate all algorithms: we run anomaly detection on each data point by feeding its historical time series values of window size $H$ into the algorithms. We skip the detection if there are less than 14 days' historical values before the detection day, or if there are any missing values within the window. We choose $H=14$ days when there are 14 to 28 days' historical data, and choose $H=28$ days when there are 28 or more historical data points. For AlerTiger, the latest observed time series is fed into the model for estimating its anomaly probability. For other algorithms, we treat a data point as abnormal if its value is outside of prediction boundaries. In the evaluation, we disable all filters and apply the same grouping logic across across all algorithms to compare the pure anomaly detection performance.

For individual algorithm: AlerTiger is trained on the "MLOps Training Dataset" using the Adam optimizer with a batch size of 256 over 100 epochs; we use a 5-layer MLP model for regression network, where the first 3 layers (layer size 18, 9, 8) are shared while the last two layers (layer size 4 and 1) are specific for baseline, lower boundary, and upper boundary. We choose anomaly score cutoff of 0.2 and quantile loss's boundary 2.5\% and 97.5\%, both parameters are tuned on the evaluation fold of training dataset. Similarly, for DeepAR, we applied the open-sourced codebase \cite{tensorflow2019deepar} to train a model on the same training dataset. For the other models, Prophet, ARIMA, AR, and SARIMAX, we use the default parameters for forecasting, any data points outside the confidence interval are considered anomalous. 

\subsection{Evaluation Metrics}
\label{subsection:Evaluation Metrics}
A straightforward approach to evaluating the classification performance is evaluating every daily data point. However, such a pointwise evaluation implicitly weighs each of the continuous anomalous durations proportionally to their respective durations, which is not the case in production where we treat each anomalous period as separate individual issues triggering a single alert. Therefore, we use an interval-wise level evaluation \cite{tatbul2018precision} by treating each continuous anomalous duration as one issue. We determine the label of each of these intervals by checking if it overlapped with any positive labels in our benchmark dataset. The final TP, FP, TN, and FN are estimated at the interval level. If an interval is longer than seven days, we chop them into intervals of 7-day max length. For example, if a three-day interval is labeled as abnormal, and an algorithm detects a two-day anomaly within the same interval, then this interval is considered one True Positive. 

\textbf{Precision, Recall, and F1 score} are calculated using the TP, FP, TN, and FN values generated using the interval-wise evaluation. We care more about the precision and F1 score more because we do not want to send a high number of false alerts to the AI teams since a large number of false alerts will make it infeasible for AI teams to invest in investigating alerts that eventually were false alerts in the first place. However, we still want to capture as many true issues as possible while not compromising on precision.

\textbf{Execution Time} is a critical aspect of our system as prompt notifications are necessary to mitigate or resolve critical issues. To achieve this, our solution needs to be lightweight, scalable, and have low computational resource requirements, making it suitable for shorter granularity or even real-time alerting. In our experiments, we evaluated the execution time of each algorithm on 8 CPU cores with 16 threads enabled for multithreading.

\subsection{Experiment Results}
\label{subsection:Experiment Results}
Based on the evaluation metric defined above, we compare our algorithm against five different algorithms on the "MLOps Evaluation Dataset" Tables \ref{tab:Model comparison without filters}. The evaluations is at the model level since the production dataset is labeled at the model level. As we can see in Table \ref{tab:Model comparison without filters}, the "AlerTiger" algorithm outperforms all other approaches on MLOps Benchmark Dataset in terms of precision, recall, F1 score, and execution time. 

\begin{table}[]
    \centering
    \begin{tabular}{l|c|c|c|c}
     \hline
         Algorithm  & Precision & Recall & F1 Score & Execution Time \\\hline
         AlerTiger & \textbf{0.47} & \textbf{1.00} & \textbf{0.64} & \textbf{0 hours 35 mins} \\
         ARIMA & 0.45 & 0.94 & 0.61 & 1 hours 28 mins \\ 
         DeepAR & 0.44 & 0.88 & 0.59 & 8 hours 49 mins \\
         Prophet & 0.44 & \textbf{1.00} & 0.61 & 14 hours 0 mins \\ 
         SARIMAX & 0.43 & 0.94 & 0.59 & 1 hours 30 mins \\ 
         AR & 0.43 & 0.94 & 0.59 & 1 hours 30 mins \\\hline
    \end{tabular}
    \caption{Model comparison without filters on MLOps evaluation data. Showing precision, recall, F1 Score, and execution time for each model on MLOps Evaluation Dataset. The best model for each column is in bold}
    \label{tab:Model comparison without filters}
\end{table}

\subsection{Ablation Test}
\label{subsection:Ablation Test}
In order to evaluate the importance of each component of our proposed method, we conduct a series of ablation studies. Starting from the AlerTiger system with anomaly filtering enabled, we removed or altered the filtering components and classification components and measured the system performance using precision, recall, and F1 score. The experiment results are in Table \ref{tab:ablation test}.

The first experiment focuses on the impact of anomaly filtering. We applied the same AlerTiger algorithm for univariate anomaly detection but removed the filters and saw the impact on the evaluation metrics. The result, presented in Table \ref{tab:ablation test} (rows 1 to 4), shows that removing the anomaly filters will reduce the alert precisions and F1 scores and trigger more false alerts. This suggests that having filters plays an important role in noise alert reduction.

The second experiment evaluates the effectiveness of  the supervised anomaly classification layer. We removed the supervised anomaly classification component and replaced it with a confidence interval classifier where out-of-boundary observations are considered abnormal. The result, presented in Table \ref{tab:ablation test} (rows 4 to 5), shows that removing the supervised anomaly classifier will reduce alert precision and F1 score. This suggests that having a supervised anomaly classifier is important.

\begin{table}[]
    \centering
    \begin{tabular}{l|c|c|c}
     \hline
         Group  & Precision & Recall & F1 Score \\\hline
         AlerTiger + All Filters & \textbf{0.50} & 0.94  & \textbf{0.65} \\
         AlerTiger + Duration Severity Filters & 0.49 & \textbf{1.00} & \textbf{0.65} \\
         AlerTiger + MTR Filter & 0.49 & 0.94 & 0.64 \\
         AlerTiger & 0.47 & \textbf{1.00} & 0.64 \\
         AlerTiger forecast only & 0.46 & \textbf{1.00} & 0.63 \\ \hline
    \end{tabular}
    \caption{AlerTiger ablation test on MLOps evaluation data. Showing precision, recall, F1 Score on MLOps Evaluation Dataset. The best model for each column is in bold}
    \label{tab:ablation test}
\end{table}

\subsection{Effect of Time Series and Anomaly on Performance}
\label{subsection: Effect of the change of hyperparameters for synthetic dataset}
\begin{figure}[hbt!]
    \centering
    \includegraphics[width=0.47\textwidth]{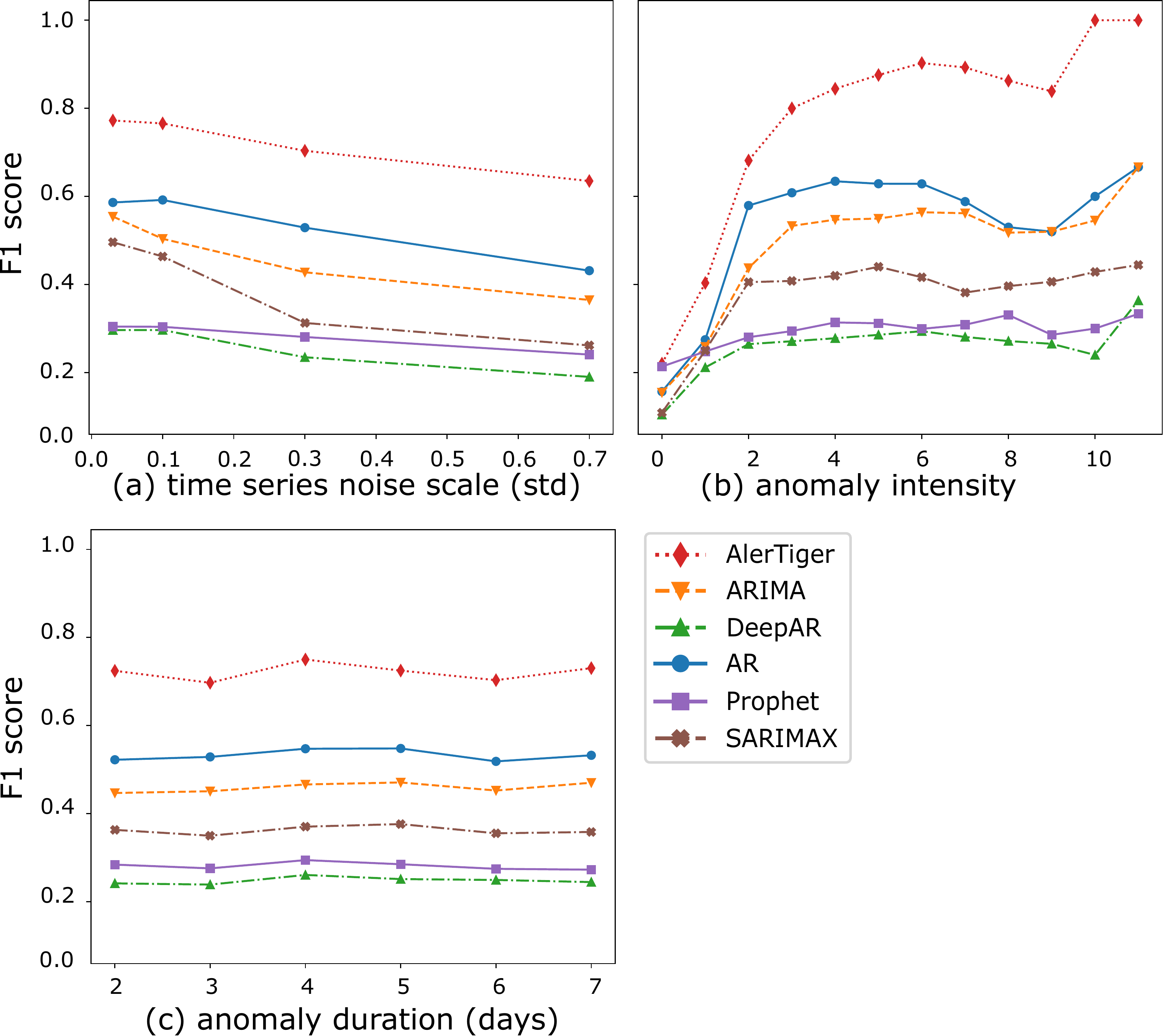}
    \caption{Anomaly detection performance on various anomalies and time series}
    \label{fig:performance evaluation on synthetic dataset}
\end{figure}

With the synthetic data, we can study how the detection performance of different algorithms varies with change in different data generation processes. We compare the F1 score of each of these algorithms with changes in time series' noise varied by the standard deviation (Figure \ref{fig:performance evaluation on synthetic dataset}.a), anomaly intensity (Figure \ref{fig:performance evaluation on synthetic dataset}.b), and anomaly duration (Figure \ref{fig:performance evaluation on synthetic dataset}.c). With an increase in the noise in the time series, the performance of each algorithm drops, clearly showing that increased noise in the time series may confuse the model to identify the anomaly effectively. As we increase in anomaly intensity, we see a rise in the F1 score for all algorithms. This shows that the model becomes more confident identifying anomalous behavior as the anomaly intensity increases. One interesting phenomenon we observed is that with an increase in anomaly duration, no drastic change occurs in the performance of each algorithm. We believe the reason is that if the model catches anomalies in the first couple of days, an increasing duration for the same intensity may not impact performance more. For each experiment, the AlerTiger algorithm clearly outperforms the other algorithms in terms of F1 score for all data types.

\section{Production Case Studies}
We have deployed the AlerTiger system on LinkedIn's ProML platform. Since then, it has captured many model issues that have led to significant business gains once fixed. We show two representative success stories here.

As one example, we identified an overtime feature distribution change across 9 features of a production model, and sent a timely alert to the model owners. Upon further investigation, the model owners found this feature distribution change was caused by a UI migration which changed the tracking data schema. This change resulted in missing observed features for affected members, leading to a decrease in model performance. The AI team retrained the model with the new schema, which achieved a significant business metric gain. 

In another example, when a new model was being ramped, we detected several features with the same values for all members, causing inconsistent online performance from the offline experiments. This concerning inconsistency blocked further model rollout. These features turned out to be among the most important features for the model. After an investigation, the model owners discovered the default values used in the online system led to this inconsistency. After fixing this issue, we ramped the new model and realized the expected business gain.

\section{Conclusion}
\label{section:Conclusion}
In this paper, we propose AlerTiger, a deep-learning-based MLOps model monitoring system that helps AI teams across the company monitor their AI models' health by detecting anomalies in models' input features and output score over time. Our solution generates three categories of statistics to indicate AI model health, offers a two-stage deep anomaly detection solution to address label sparsity and attain the generalizability of monitoring new models, and provides holistic reports for actionable alerts. Our system has high precision and recall in detecting anomalies with fast execution time. The success of our approach is demonstrated by its implementation across most of LinkedIn's production AI models for over a year, resulting in numerous improvements to AI models.

\begin{acks}
This work is done by the AI Quality Foundation team at LinkedIn. Special thanks to our  collaborators in ProML Health Assurance Platform Team, Ion Team on infrastructure support and development, and LinkedIn's AI teams for early adoption and for sharing constructive feedback. In particular, thanks to Niranjan Balasubramanian, Senthil Mani, Vikram Singh, Vipul Mathur, Mohit Gupta, Pratik Patre, Radhika Sharma, Shubham Gupta, Tanaya Jha, Akshay Uppal, Priyanka Saxena, Ankit Dhankhar, Bala Rajan, Leo Sun, Bryan Chen, Curtis Wang, Rachit Arora, Jihao Zhang for the infrastructure support and development. Thanks to Swathi Varambally, Shreya Mukhopadhyay, Girish Doddaballapur for technical product management. Thanks to Albert Chen and Reza Hosseini for reviewing and providing guidance on the paper. We also thank the leadership team of Zheng Li, Souvik Ghosh, and Ya Xu for their support.
\end{acks}

\bibliographystyle{ACM-Reference-Format}
\balance
\bibliography{main}

\end{document}